# An approach to automated videogame beta testing

Jennifer Hernández Bécares, Luis Costero Valero, Pedro Pablo Gómez Martín *

*Facultad de Informática, Universidad Complutense de Madrid, C/ Prof. José García Santesmases, 9, 28040 Madrid, Spain*

ABSTRACT

Videogames developed in the 1970s and 1980s were modest programs created in a couple of months by a single person, who played the roles of designer, artist and programmer. Since then, videogames have evolved to become a multi-million dollar industry. Today, AAA game development involves hundreds of people working together over several years. Management and engineering requirements have changed at the same pace. Although many of the processes have been adapted over time, this is not quite true for quality assurance tasks, which are still done mainly manually by human beta testers due to the specific peculiarities of videogames. This paper presents an approach to automate this beta testing.

## 1. Introduction

Today, videogames constitute an industry with a revenue comparable with the one in the motion picture industry. As an example, *Grand Theft Auto V*, from Rockstar Games, generated $800 million worldwide during its first 24 h on sale in 2013 [1].

Such a success does not come cheap. Videogame development involves a huge amount of high-skilled people with very different roles, from programmers to designers, artists and composers to name just a few. They must work together in long productions that can last several years. Again, the development of *Grand Theft Auto V* required 5 years and up to 1000 people working in seven different locations [2].

Quality assurance in such big software artifacts is a huge challenge. Although classic techniques such as *unit testing* can still be used, videogames have some peculiarities that usually require manual testing, which increase the already high costs. For example, videogames push *hardware* to the limits, but the user experience must be adequate in mid-range PCs and, at the same time, make high-level gaming PCs worthy. *Compatibility tests* become a nightmare due to hidden *hardware* relationships that become more and more complex over time.

On top of that, videogames are not just software. A big percentage of the development budget is devoted to the so-called *assets*, which define how the game look (3D models, textures, music, etc.) and the general gameplay (maps, missions and puzzles). Each time a level is changed or finetuned, gameplay errors might be introduced, preventing players from completing it. Although these problems cannot be considered *bugs* (at least not *software* bugs), they ruin the game, so they must be detected and solved.

In this paper we propose a way of automatising *videogame beta testing*, useful for testing the game not only after making changes in the source code but also for proving that the playability of a game and the global gameplay are still correct after introducing *level changes*. Next section covers some related work, and Section 3 describes component-based architectures, which have become the standard for videogames in the last decade. Section 4 introduces videogame testing and its limitations, whilst Section 5 talks about our proposal of carrying out automatic beta tests for videogames. Sections 6 and 7 introduce Petri Nets and how to model a game and run tests using them. After that, Section 8 puts into practice the ideas explained in previous sections in a small videogame, and Section 9 describes some implementation details. Finally, this paper ends with conclusions and future work.

## 2. Related work

With systems growth in size and complexity, tests are more difficult to design and develop. Testing all the functions of a program becomes a challenging task. One of the clearest examples of this is the development of online multiplayer games [3]. The massive number of players make it impossible to predict and detect all the bugs. Online games are also difficult to debug because of the non-determinism and multi-process. Errors are hard to reproduce, so automated testing is a strong tool which increases the chance of finding errors and also improves developers efficiency.

*Monkey testing* is a black-box testing aimed at applications with graphical user interfaces that has become popular due to its inclusion in the Android Development Kit.[1] It is based on the theoretical idea that a monkey randomly using a typewriter would eventually type out all of the Shakespeare's writings. When applied to testing, it consists on a random stream of input events that are injected into the application in order to make it crash. Even though this testing technique blindly executes the game without any particular goals, it is useful for detecting hidden bugs. This technique can be improved if logs are analysed after each *monkey test* and an evolutionary algorithm is fed with the conclusions in order to make the test more and more destructive [4].

* Corresponding author.
*E-mail addresses:* jennhern@ucm.es (J. Hernández Bécares), lcostero@ucm.es (L. Costero Valero), pedrop@fdi.ucm.es (P.P. Gómez Martín).

Due to the enormous market segmentation, again specially in the Android market but more and more also in the iOS ecosystem, automated tests are essential in order to check the application in many different physical devices. In the cloud era, this has become a service provided by companies devoted to offer cloud-based development environments. For applications with graphical user interfaces, this testing based on *test cases* requires specific frameworks that check whether the GUI meets its specifications. Examples of such a software are Selenium (for web applications) and Appium (for Android and iOS). They must cope with GUI changes whilst maintaining the original *test cases* still valid, one of the problems that we address in this paper for the videogames field.

In any case, unfortunately, all those testing approaches are aimed at *software*, ignoring the fact that games are also maps, levels and puzzles. Application-level tests usually need to be adapted (or even completely recreated), even when levels suffer small changes. We are not aware of any approach to carry out automatic beta testing that pursues solving this issue as we describe in this paper.

On the other hand, researchers have been trying for a long time to create intelligent systems that can learn by demonstration how to play a videogame using traces generated by expert players. This is specially useful for those domains where creating a plausible artificial intelligence is complex, as in real-time strategy games.

Once the traces have been analysed offline, those systems play the game replicating or modifying the expert movements. Sometimes, an initial step is required, where experts annotate the traces to incorporate extra useful domain knowledge.

A good introduction to this subject is provided in [5], where efforts are described to design a generic recording system with the purpose of annotating this traces later easily. Ontañón et al. [6] shows how to adapt traces recorded in order to be able to replay them afterwards. It also shows a description of how to detect goals and subgoals of the traces recorded and how to use Petri nets for modifying the traces.

Previous research in this *learning from demonstration* field was our source of inspiration for the automatic beta testing presented in this paper.

Next section introduces both *hierarchy architectures* and *component-based architectures*, and also their differences and the basics of *message passing*, which highlights why using the second type is better and useful for us in our purpose of running automated tests.

## 3. Game architecture

Videogame development constitutes a big challenge today. Broadly speaking, there are two different aspects that must be covered. The first one refers to the *technological requirements*, including graphics, sound, physics simulation or network communication, to name just a few. The second one is related to the *game* itself, the playful characteristics that the software must provide in order to be enjoyable. This is usually known as *gameplay* or *game mechanics* and it is built with *interactive elements*.

Many of the main runtime components of a modern videogame belong to the *technological requirements* and provide the basic infrastructure needed for creating the game interactive simulation. Only a small part is designed with a particular game in mind.

Apart from those runtime software elements, videogames need *resources* such as 3D models, textures or sounds for providing the interactive experience. Early videogames had all those resources hard-coded, but today no one conceives resources in that form and they are virtually always provided using external files. They are collectively known as *assets*.

The separation between source code and *assets* constitutes the key point of the *data-driven architectures* that allow software reusability: the game aspect can be completely changed without involving programmers.

But the videogame would continue to be exactly the same if the logic or game rules were hard-coded into the source code, preventing software from being reused to create *different* games. This can be solved if the game mechanics become assets themselves, or if they are correctly isolated in the source code to be easily replaceable. The term *game engine* is used to refer to "software that is extensible and can be used as the foundation for many different games without major modifications" [7]. Ideally, all the gameplay would be specified throughout assets using external files, and the game engine would be completely independent from the game itself. Usually, game engines (such as Unity3D or Unreal Engine) include *tools* used to create all the assets that the engine will load and run.

What distinguishes one game from another are the *game mechanics*, which emerge from game interactive elements such as the player avatar abilities, the non-player characters (NPCs) or any other element of the game logic with a certain behaviour. Those interactive elements are known as *entities* or *game objects* and they are responsible for creating the game experience. Essentially, a game engine is an *entities manager*, and those entities invoke the underlying subsystems to indicate the way they must be drawn, when and how they should generate a sound, or how they should react to specific events.

The classical way of programming the *entities* is using inheritance. An abstract class is used as the base class for all the other entity classes, and the game engine stores a list of instances of this base class. In the main game loop, all the entities are updated (using an abstract method) so they have the opportunity to react to events and modify their internal state. Afterwards they are drawn using a second abstract method. Entities' behaviour depends on the concrete implementation of those abstract methods.

Depending on the game, the hierarchy can have multiple levels, with intermediate classes that provide auxiliary methods used from different subclasses. As an example, the old 1998 game, *Half-Life* [8], had an entity hierarchy composed of 9 abstract classes and 10 concrete classes shown in Fig. 1.

Although they were extensively used during the nineties and early noughties, today entity hierarchies have fallen into disuse. With videogames getting bigger and bigger, more features were added to entities and many scalability problems arose. Specifically, programmers had to decide how to split a class at each level, but entities have different facets, depending, for example, on their game logic, graphics representations or physics behaviour. Class hierarchies are static, and making the wrong choice on how to split a class in different levels can be fatal if designers suggest to add new entities later. This situation can lead to problems such as *multiple inheritance*, which is not supported in many object-oriented languages. Even if it was, we would most likely suffer from the *diamond problem*.

[1] http://developer.android.com/tools/help/monkey.html.

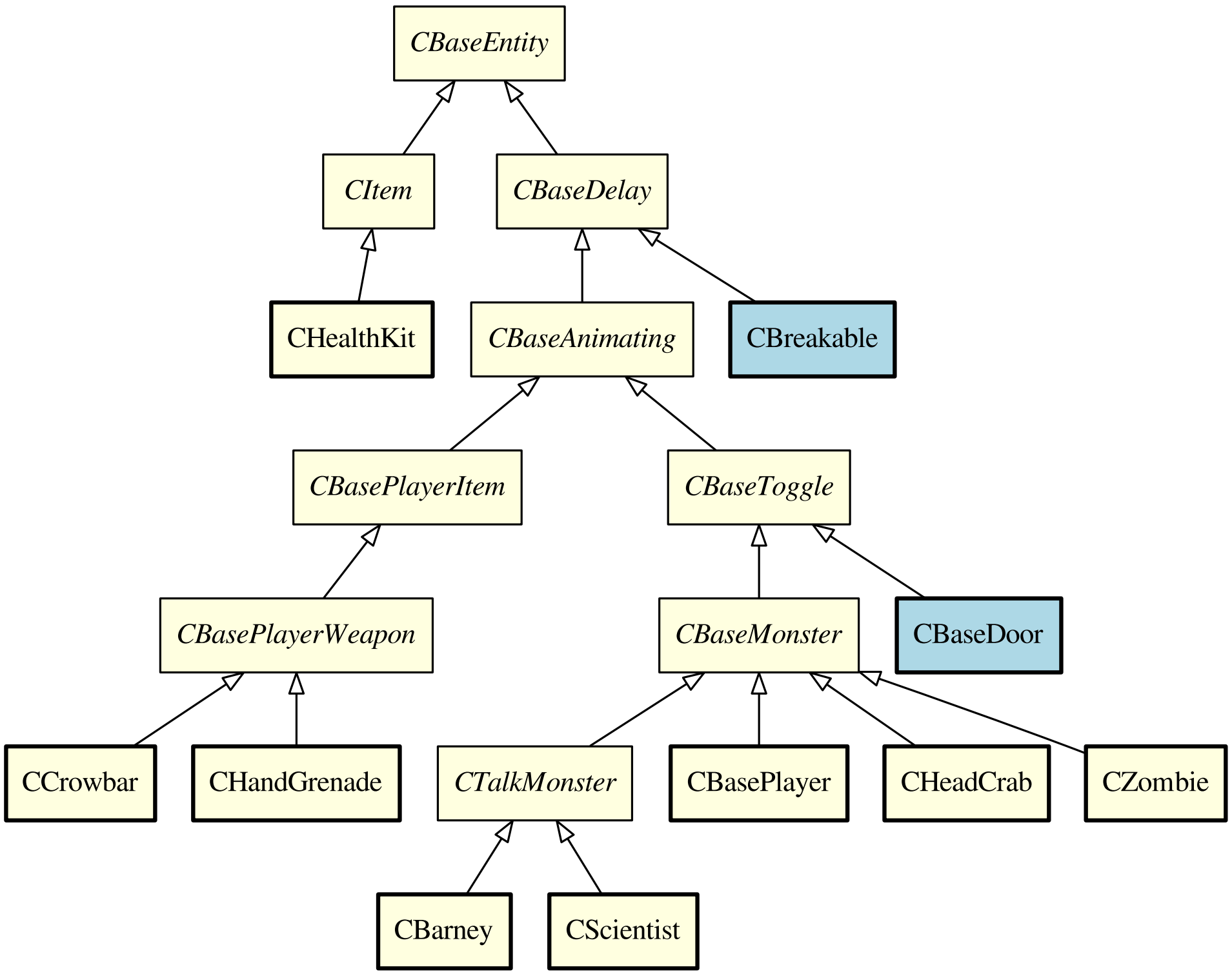

**Fig. 1.** Half-Life entities hierarchy.

But that is not the only problem of using inheritance [7]. Hierarchies are very *hard to create*, and once they are created, they are *hard to change*, since they are static. Moreover, wide hierarchies are *difficult to understand* and *difficult to maintain*. When trying to add functionalities to the entities, it can be inevitable to suffer the *bubble-up effect*, which consists on moving methods from one class to some of its predecessors to share code needed by some of the unrelated classes.

The most usual way to solve these problems is to convert the *isA* relationships into *hasA*, swapping class inheritance for composition [9,10].

Although a breakable door *is* both a door and a breakable entity, it can be also considered as a plain entity that *has* the ability to be opened and closed, and, at the same time, can be broken. Entities become *containers* of attributes (data) and *behaviours* (logic), which are coded using *components*. Components provide independent features to the entities, such as the ability to be drawn, to simulate physics, or to be damaged. Only an entity class exists, and it is composed of a list of components. Fig. 2 shows an excerpt of a possible component hierarchy for implementing Half-Life 1. Note that subclasses from `CBaseEntity` have vanished. What previously would have been a `CHandGrenade` is now a `CBaseEntity` containing an instance of a `CStaticGraphicComponent` and a `CPhysicsComponent`, amongst others.

Entity features are now decided during runtime using aggregation and statically typed entities does not exist anymore. However, now we have entities as a set of attributes and behaviours that resemble the so-called *duck typing*. The former entity subclasses are now *data* that relates a name ("Door", "Health kit") with the components it is composed of. This information is usually known as *blueprints*.

Fig. 3 shows a possible relationship between the old concrete classes and the new components, using a common grid layout. Our hypothetical breakable door would be composed of the same components of a door, plus `CDamageComponent`. Even better, this addition could be done without source code changes.

Note that we could change the aggregation model for each type of entity depending on the circumstances. For example, we could create all our entities without the graphical components for headless server executions.

Components resemble the strategy pattern [11], although they include also a *state*. When used, games are said to follow a *component based architecture*. Components are used to isolate each entity facet into a different class, focusing on them separately. This eases implementation and the use of Commercial Off-The-Shelf (COTS) modules [12].

Although it solves the problems suffered by the former entities hierarchies, the use of components have the added difficulty of communication. In order to be useful, components should *not* have hard-coded dependencies between them, so, in particular, they should not invoke methods between them. Instead of that, a generic way of broadcasting information and requests must be used.

This is usually achieved through *message passing*. When a component wants to communicate any event, for example that an entity has just been killed, it encapsulates this information into a *message* that is broadcasted to all the other components in the entity. They will decide, in turns, if the message is interesting or should be discarded. For example, the graphical component could consider to run a specific death animation.

Message passing is general enough to be used also for communication between entities. These messages are essential when designing the playability of the game. For example, a *bullet* could notify the target entity that it has been injured, triggering the previous example. Or the player entity could notify a *crowbar* that it has been used.

As a conclusion, messages can be seen as the declarative representation of a method invocation. Although this has a performance penalty, it is also crucial for the testing techniques proposed in this paper.

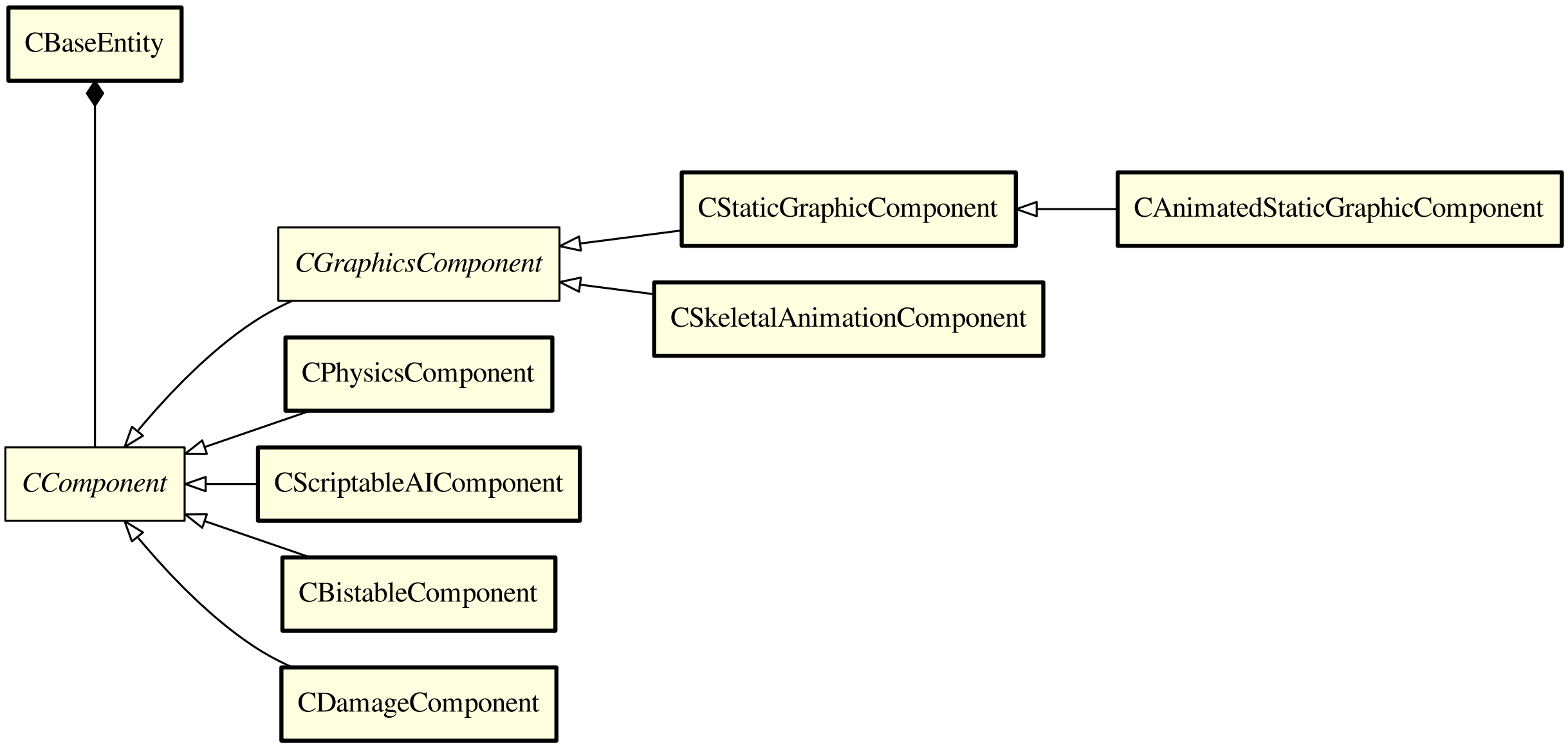

**Fig. 2.** Hypothetic component hierarchy excerpt for Half-Life 1.

## 4. Videogame testing and continuous integration

Testing is defined by the IEEE Computer Society [13] as the process of analysing a software item to detect the differences between existing and required conditions and to evaluate the features of the software item. In other words, testing is a formal technique used to check and prove whether a certain developed software meets its quality, functional and reliability requirements and specifications. There are many testing approaches, each one designed for checking different aspects of the software. For example, a test can be done with the purpose of checking whether the software can run in machines with different hardware (*compatibility tests*), or whether it is still behaving properly after a big change in the implementation (*regression tests*). Alternatively, expert users can test the software in an early phase of the development (*alpha or beta tests*) to report further errors.

*Unit testing* is a particularly popular test type designed to test the functionality of specific sections of code, to ensure that they are working as expected. A unit test is a short code fragment written with the sole purpose of checking that another piece of code (*unit*) works as expected. It will not be part of the final product, but it will be run occasionally against the source code base in order to validate it. When a certain software item or software feature fulfils the imposed requirements specified in the test plan, the associated unit test is passed. Pass or fail criteria are decision rules used to determine whether the software item or feature passes or fails a test. Passing a test not only leads to the correctness of the affected module, but it also provides remarkable benefits such as early detection of problems, easy refactoring of the code and simplicity of integration. Detecting problems and bugs early in the software development lifecycle translates into decreasing costs, whilst unit tests make possible to check individual parts of a program before blending all the modules into a bigger program.

Unit tests should have certain attributes in order to be good and maintainable. Here we list some of them, which are further explained in [14, Chapter 3]:

- **Tests should help to improve quality**.
- **Tests should be easy to run**: they must be fully automated, self-checking and repeatable, and also independent from other tests. Tests should be run with almost no additional effort and designed in a way that they can be repeated multiple times with the exact same results.
- **Tests should be easy to write and maintain**: test overlap must be reduced to a minimum. That way, if one test changes, the rest of them should not be affected.
- **Tests should require minimal maintenance as the system evolves around them**: automated tests should make change easier, not more difficult to achieve.

Unit testing splits the program into modules (units) and checks whether each individual part is correct. As a general testing technique, it is suitable for game development to check the robustness of independent modules of a general architecture. It is useful to check, for example, the math library, the collision detector, or resource loaders.

Unit testing has gained wide acceptance between developers thanks to the existence of unit testing frameworks, which eases unit test creation and execution. Running a big set of tests becomes a child's play, so they can be checked often, reassuring and giving confidence to developers when they change the code base and helping the continuous integration process.

Unfortunately, unit testing has limitations. It cannot be used to detect *integration errors*, to check software performance or to test multithread programs, as games usually are. Even worse, traditional software tests are not enough to check all the features that a game has. Things such as playability and user experience need to be checked by *beta testers*, who are human users that play the game levels over and over again, doing something slightly different every time. Their purpose is to find bugs, glitches in the images or incorrect and unexpected behaviours. They are part of the Software Quality Assurance Testing Phase, which is an important part of the entire software development process. Using beta testers requires a lot of time and effort, and increases development costs. In fact, testing is so important and expensive that has become a business by itself, with companies earning millions of dollars each year and successfully trading on the stock market [15].

This paper proposes a theoretical framework and a proof of concept for bringing the unit test automation into game beta testing. The main objective is to be able to run the game over and over again with no human intervention after changing the source code and the assets.

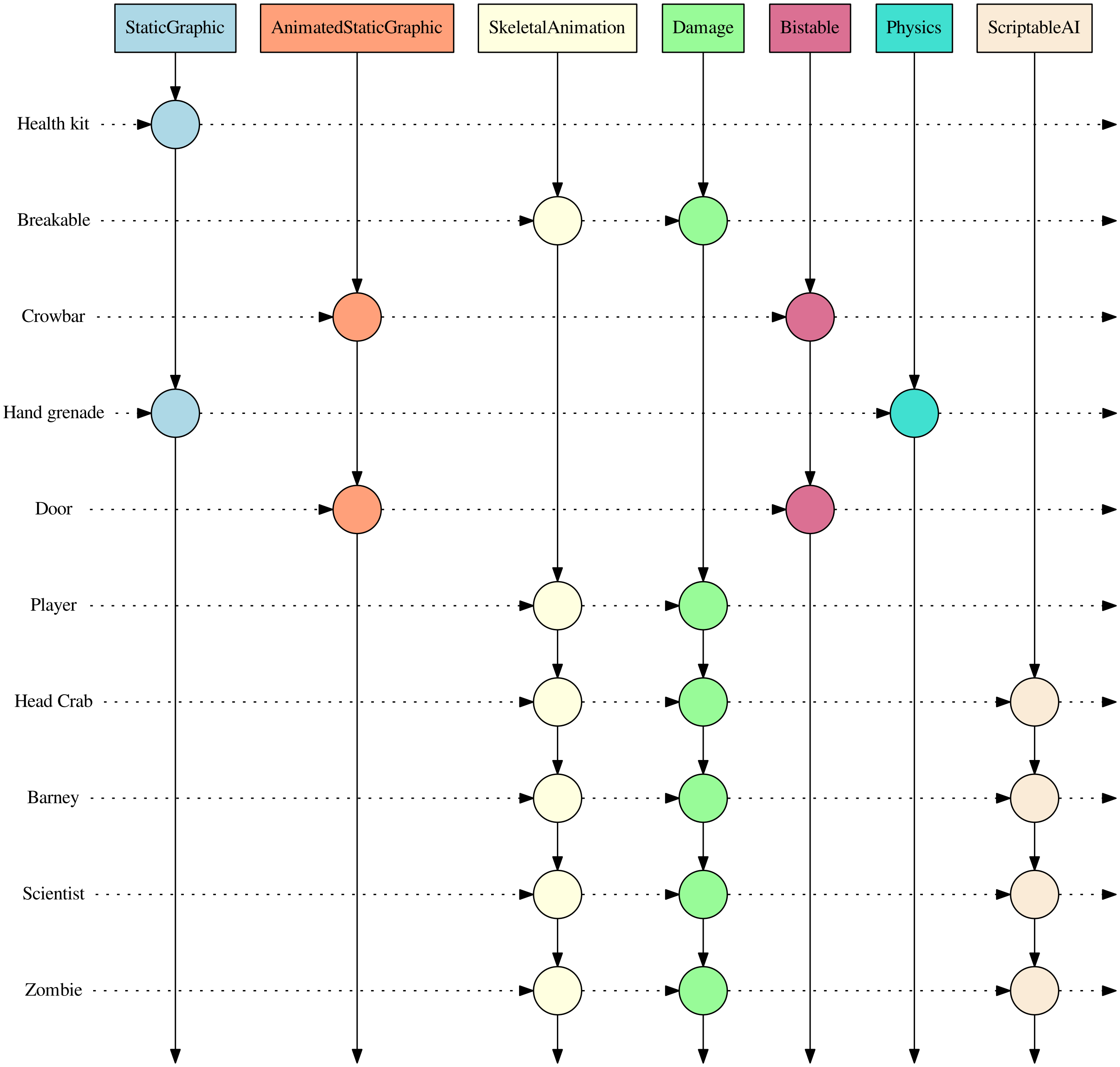

**Fig. 3.** Hypothetic component grid for Half-Life 1.

## 5. Automatic beta testing

### *5.1. Recording game sessions for compatibility and regression tests*

Nowadays, a lot of development teams are recording game sessions when beta testers play the game, saving both keyboard and mouse input logs. These logs are an invaluable resource if the game crashes because they allow to replay the game session in a development computer exactly as it was done by the beta tester. Keep in mind that beta testers' computers are usually quite different from those used by programmers and, in fact, the game version used by them is also different, without all the debug information embedded. In order to be able to re-execute the game session, some implementation peculiarities must be kept in mind, as described in Section 9.

Input logs are also useful in not so dramatic scenarios, for checking what went wrong if there were bugs or glitches caught up by a beta tester. Programmers will use the logs to replicate the errors themselves.

We propose the use of those input logs as a first step towards automatic gameplay testing, based on the idea that using human beta testers increases costs and requires a huge amount of time and effort. Because of this, it is profitable to avoid hiring beta testers to play the same levels over and over again every time developers introduce changes in the game. To improve the process of testing, our suggestion is to use the recorded games to replay the session automatically without any human intervention.

Our use of those logs is now completely different. Nowadays, it is common to replay game sessions that *went wrong* in order to replicate bugs. Our idea is to use the logs that did *not* provoke errors, but made the player end a level, for example, to test that everything is still working correctly. They are, therefore, used for running *compatibility* and *regression tests*. Using logs of successful plays allows us to run game sessions under different hardware configurations or after software changes. Thus, we can inject keyboard and mouse inputs to easily simulate beta testers' executions to check if everything is still working properly after all the modifications, no matter whether these changes were hardware or software-based.

In the next subsection we will propose a way of using these ideas to improve the feedback received to help programmers and developers understand what works and what does not work.

### 5.2. Enhancing automated tests: improving feedback

The output of the compatibility and regression tests mentioned in the previous subsection is a yes/no answer that represents for example if the level was completed successfully or not. This is useful to avoid almost all human interaction in the testing process when the execution of the game goes right, but we can do it better when the execution fails. Specifically, we are interested in providing richer feedback regarding the causes that made the game fail.

Initially, during the recording session, we are only storing the keystrokes and mouse movements carried out by the human player. The game will be simulating the environment that creates the interactive experience. Underneath, that means that the system knows, for example, the exact moment when an entity died or the player used a crowbar. If the game uses a component based architecture with message passing (Section 3), all those events are triggered by messages, as a declarative way of method invokation. The messages broadcasted during a game session constitute a complete *internal trace* that describe what happened to all the entities during the gameplay.

This information is priceless for automatic testing. Instead of just storing the user input, we can also save those generated messages. Many of them will be unimportant and noisy (for example, each entity delta movement) so we can filter them out and only gather those related to "high-level actions" that make the gameplay advance.

When replaying the log, it is not difficult to compare the initial log and the new log generated by the automatic test and report all the differences. Thus, specific errors can be detected and notified. For example, if we were expecting the player to shoot an enemy but the collision between the bullet and the character is not reported as expected, the test can be stopped and a problem in the physics engine will be reported. Afterwards, developers can look into it and try to figure out why the physics engine is causing this behaviour.

Note that this message tracking will be done every time the test is executed. Although a test could have a valid global result (for example, "the automatic player ended the level"), some discrepancies could have occurred at the messages level. This is a signal showing that something has changed. Maybe it is not enough to modify the outcome of the test, but sufficient to provoke internal changes. The information contained in these messages is also useful for developers, who should keep an eye on them.

### 5.3. Enhancing automated tests: adaptation

Running automated tests using the traces recorded when human betatesters play the game is, as highlighted in the previous section, very helpful for automatic compatibility and regression tests. They are useful to confirm that the game continues to work correctly after code changes due to bug fixing, or after code refactorization.

However, videogames are composed not only of source code but also of assets (Section 3). Map designers could decide to modify a level by moving entities, adding new elements or simply by changing its static structure. Even under those circumstances, we would like to be capable of executing our old tests without generating them again by betatesters.

In order to reuse these tests, we need to come up with a different way of running them. The previous raw game replay is not suitable when the map level has been changed, since using *blind* input injection will make the player wander incorrectly in search of places and entities that may have been moved, modified or even removed from the game. Our proposal is to use the "high-level actions" described in Section 5.2 in a new way, not only to test if the original actions occur in the replay of the game, but also to manipulate the automatic player in such a way that we can make him carry out these high-level actions. For example, imagine that in the original game there was a button (let's call it B1) located in $(0.0, 0.0, 0.0)$. In this case, the high-level action is not associated to that particular position, but to find B1 in the map.

The previous idea works because those actions hold the gameplay intentions of the player, which will lead to the completion of the level. If the map has been slightly modified, replaying actions like this can ensure that the changes introduced have not broken it and that the gameplay remains the same. This is what we call "*high-level unit tests*".

To create them, our approach is to record game sessions and then execute the same *high-level actions* again. This is different to the previous automated tests because instead of injecting input events such as "spacebar, left, left, spacebar, right, return" we are going to reproduce the real actions recorded, which represent the result of replaying a list of keystrokes and mouse movements. For example, the sequence described can represent how the user carried out the action "push button A". However, if the level designer modifies the map and button A is no longer in the same position, injecting those events is useless and hence we need to build a system that understands how to carry out the action "push button A", no matter where the button actually is.

What our system does is to try to replicate the state of the game when the action took place before attempting to perform the action. If the stored log contains all the necessary attributes of the game, replicating the state should not be problematic. To replay an action in this way, it is required to know which entity performed the actions, what action took place, the target entity to these actions (if any) and the time frame to complete the action. As an example, imagine an action in which "Player 1" wants to grab an object from the ground. In this case, the action is "Pick object", the entity performing the action is "Player 1" and the target entity is the object. In this situation, knowing the time frame is vital, considering that a player cannot pick an object if the object is not there or if the object is not close enough to reach it. Therefore, the state of the game needs to be as close as possible as the state logged when the human tester played to make the replay of the action feasible.

On the other hand, storing the position of the objects in the map in the log is not required, as that information is already stored in a map file and it can be loaded into the game and accessed in the runtime. Moreover, if we were to store positions and they were changed in the map file afterwards, we could be replaying actions in the wrong way after any changes in the map.

In the following section we will discuss the kind of tests that were run by using the stored logs of beta tester executions.

## 6. Replaying traces and running tests

In Section 5 we proposed some ideas needed to perform automatic beta testing. The main idea of the system is to launch the game in a specific way, especially implemented for testing purposes, allowing us to record both the raw user input (keystrokes and mouse movements) and the internal messages generated by the components. To record this information, two files are generated. Implementation requirements will be shown later in Section 9.

These two files (one containing the recorded raw input and the other one containing the internal messages) are used in two different ways, depending on the objective of the test. The first possible objective of a test is to repeat the execution exactly as the user did it, injecting the raw input and checking if the messages generated during the replay are the same as the ones recorded before. For this test, the file that contains the recorded raw input is used. This kind of execution allows us to make regression and compatibility tests providing useful feedback by comparing the internal messages; however, even the slightest modification of the map makes the raw input files invalid and they have to be recreated by playing the level again.

The second test we can run is what we called a "high-level test", where the raw user input is ignored. For this kind of test, we use the internal messages recorded previously to try to replicate the actions of the player. Again, we want to check if the new messages generated are the same as the high level messages stored before. However, contrary to the tests that use the raw user input, this one can be executed even when a level has been modified or software changes were added, finishing successfully, as described in Sections 5.3 and 7. In addition, we can use this type of test to check that the playability is still correct after all those changes.

As in unit testing frameworks, to make these tests substantially useful it is necessary to automatise their execution. Making them dependent on parameters is also a good approach. In unit testing, a *test* is a piece of code designed for checking a specific module, configuring it with an initial state, running it in a specific way and finally checking its output. As opposed to unit tests, our "automatic beta tests" are defined by a previous trace recorded before, complemented with information such as the level in which it was recorded initially, and the conditions for considering the test to be successful.

In the case of simple tests that only inject the raw input without adapting it, the simplest thing we can check is that every internal message recorded before was generated again when expected. However, during the development phase of the game, code may have changed, making some of the original messages change or even disappear completely. In this case, comparing the new messages with the ones generated before is not valid anymore, and the programmer should filter out the information received from the test. We might even have a situation where we are only interested in the last message generated, the one notifying that the level has been completed successfully. If that is the case, we simply would like to indicate that whenever this message is generated, the test has to report a success.

Apart from indicating what messages will make a test succeed, indicating which messages should not appear during the execution of the test is helpful for the programmer. These messages are what we called immediate *failure conditions*. If one of them appears, it means that the test has failed and it is not necessary to continue executing it, saving lots of resources and time. Also, it is usual to set a limit of time to finish the execution of the tests, because some of them will never be completed and we do not want to be waiting indefinitely for a success message.

For instance, suppose a level in which we have the player character, an enemy and a button, and where the goal of the level is that the player touches the button before its enemy does. The programmer could design a test in which the success condition is that the player touches the button and the failure condition is that the button is touched by the enemy instead. Fig. 4 shows an example of this test. However, it is possible that none of them ever touch the button. Because of this, the test has an attribute called "maximum time", which will cause the execution of the game to abort, reporting a failure if that time is exceeded.

Finally, up until now we have considered the tests that use the raw input and the "high-level tests" as two different types of tests. Obviously, tests that use the raw input consume less resources, but they are less versatile when the map level has changed. To solve this situation, we can have a mixed approach where both ways of running tests can be combined into a more complex reproduction of traces where we start using the actions contained in the raw input file and then, in the event of a discrepancy between the output and the messages in the original file, the system can start to use the high-level traces to fulfil the test requirements and make it pass. This approach allows the programmer to run the high-level test without stopping and relaunching the raw input one.

Section 7 introduces Petri Nets, which have been used as a model to represent the sequence of actions carried out in the level by the beta testers when it was first played.

## 7. Modelling the game with Timed Petri nets

Petri nets [16–18] are a modelling language that can be described both graphically and mathematically. The graphical description is represented as a directed graph composed by nodes, bars (or squares), arcs and tokens. Elements of a Petri net model (shown in Fig. 5) are the following:

- **Places**: they are symbolised by nodes. Places are passive elements in the Petri net and they represent conditions.
- **Transitions**: bars or squares represent transitions, which are the actions or events that can cause a Petri net place to change. Thus, they are active elements.
- **Tokens**: each of the elements that can fire a transition are called tokens. A discrete number of tokens (represented by marks) can be contained in each place. Together with places they model system states.
- **Arcs**: places and transitions are connected by arcs. An arc can connect a place to a transition or the other way round, but they can never go between places or between transitions.

A Petri net changes from one state to the next state when a transition "fires". When this happens, tokens present in *input places* are consumed, and new ones are created in *output places*. A transition can fire when there is at least one token on each of the transition's input places. Plus, when a transition fires, one token from each of its input places is consumed and a single token on each of its output places is produced. Whenever a transition is able to fire, it is said that the transition is enabled. Otherwise, the transition is disabled.

Because several transitions can fire at the same time, Petri nets are able to *model systems with non-deterministic behaviour*, which is useful for modelling the behaviour of videogames. That is the reason why Petri nets are different from finite state machines, because they allow us to model multiple activities. In a finite state machine, there is only a single "current" state that determines which action can occur next. On the contrary, Petri nets can have several states and all of them can evolve and change the state of the Petri net. Having multiple tokens in different places allows us to fire as many transitions at the same time as tokens we have in our Petri net. Thus, multiple states can evolve in parallel causing several independent changes to occur at once in the model.

Fig. 6 shows an example of a Petri net model in a situation where the main player wants to open a door and has to push a

```
{"traces_file": "...",
 "max_time": 15000,

 "success_conditions":[
     {"type": "ordered",
      "msg" : [
        { "type" : "TOUCHED",
          "SourceEntity" : "Player",
          "TargetEntity" : "doorTrigger1"
        } ]
     }
 ],

 "failure_conditions":[
     {"type": "ordered",
      "msg" : [
        { "type" : "TOUCHED",
          "SourceEntity" : "Enemy",
          "TargetEntity" : "doorTrigger1"
        } ]
     } ]
}
```

**Fig. 4.** Example of a *high-level* unit test.

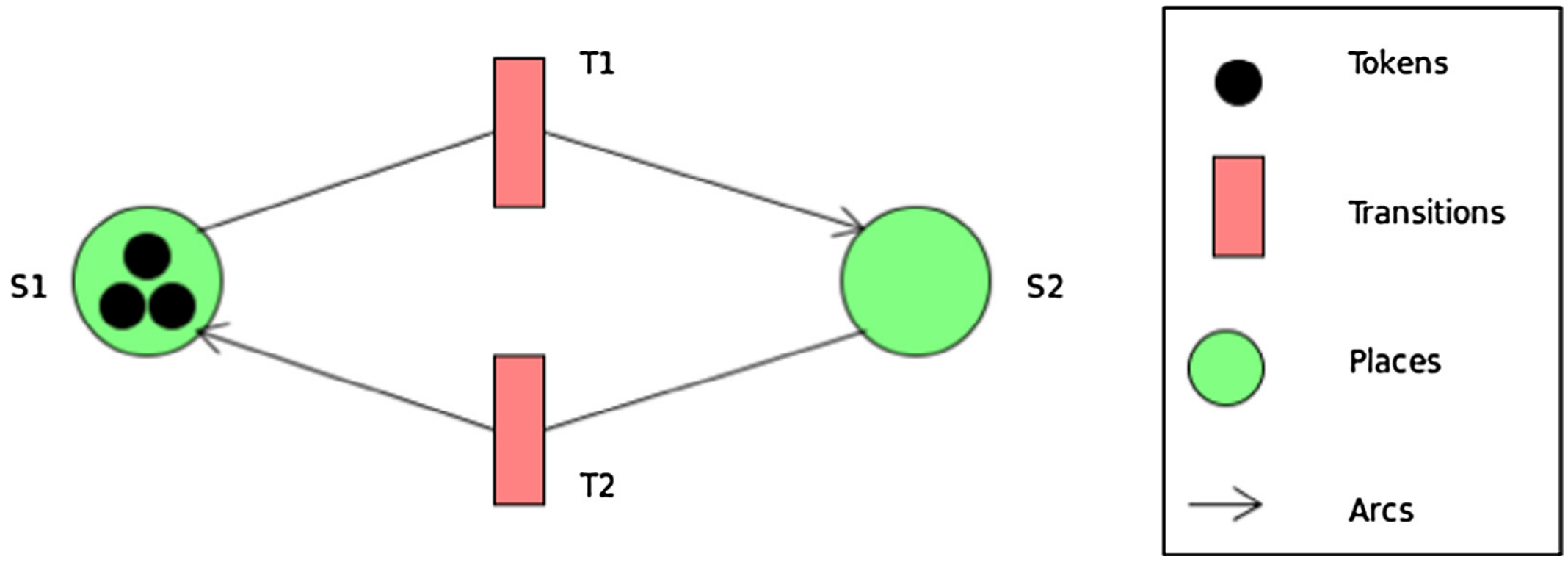

**Fig. 5.** Elements of a Petri net model.

button to open it. Fig. 6a represents the initial state, where the main player arrived at the door and the door was closed. Fig. 6b shows the Petri net model when the player already pushed the door button to open the door. Now, we have a token both in S1 and S2. Having a token in S1 means that the door is closed, and having a token in S2 means that the player pushed the button. At the point, we have a token in both of the input places of transition T1. Because of this, the transition is now enabled and it can be fired. Both of the tokens in its input places are consumed and a new single token is created in S3, as shown in Fig. 6c. The final place S3 represents that the door is now open, and the transition T1 is disabled again (an open door cannot be opened).

Continuing with our example of opening a door, it is necessary to highlight that the transition from 6b and c is not immediate. Therefore, we cannot consider this transition to be atomic with regard to the entire Petri net, since other transitions may be being fired at the same time in some other parts of the net. This lack of immediacy can be modelled thanks to the existence of *Timed Petri nets*. Timed Petri nets are an extension of classic Petri nets which allow us to introduce an associated time to each transition.

Another useful extension are *Coloured Petri Nets* [19]. Coloured Petri Nets are just a way of adding a value or type to the tokens in the Petri Net, making them distinguishable. Token values are traditionally referred as *token colours*. The different colours are useful because depending on their value, transitions will be fired or not, only moving tokens to a different place if any of the transition allowed types equals the token type. In our example, this would be useful if the door could only be opened if a specific kind of entity had pressed the button. The colour would represent the entity type, and the transition would only be fired when the token had a specific colour.

The example in Fig. 6 shows that Petri nets are a valid mechanism to model a videogame [17]. Transitions constitute changes in the general state of the game that, from the point of view of the implementation, are produced as a result of the treatment of high-level messages in the component-based architecture, such as "Open door" or "Shoot". At the same time, Petri net places and the existence of tokens on them model *preconditions* that must be fulfiled so that messages can be generated in the first place. For example, to generate a message "Open door", the door must be closed previously and a button that opens the door must be enabled; or, in order to produce a message "Shoot", the player must have a weapon and ammunition in the inventory.

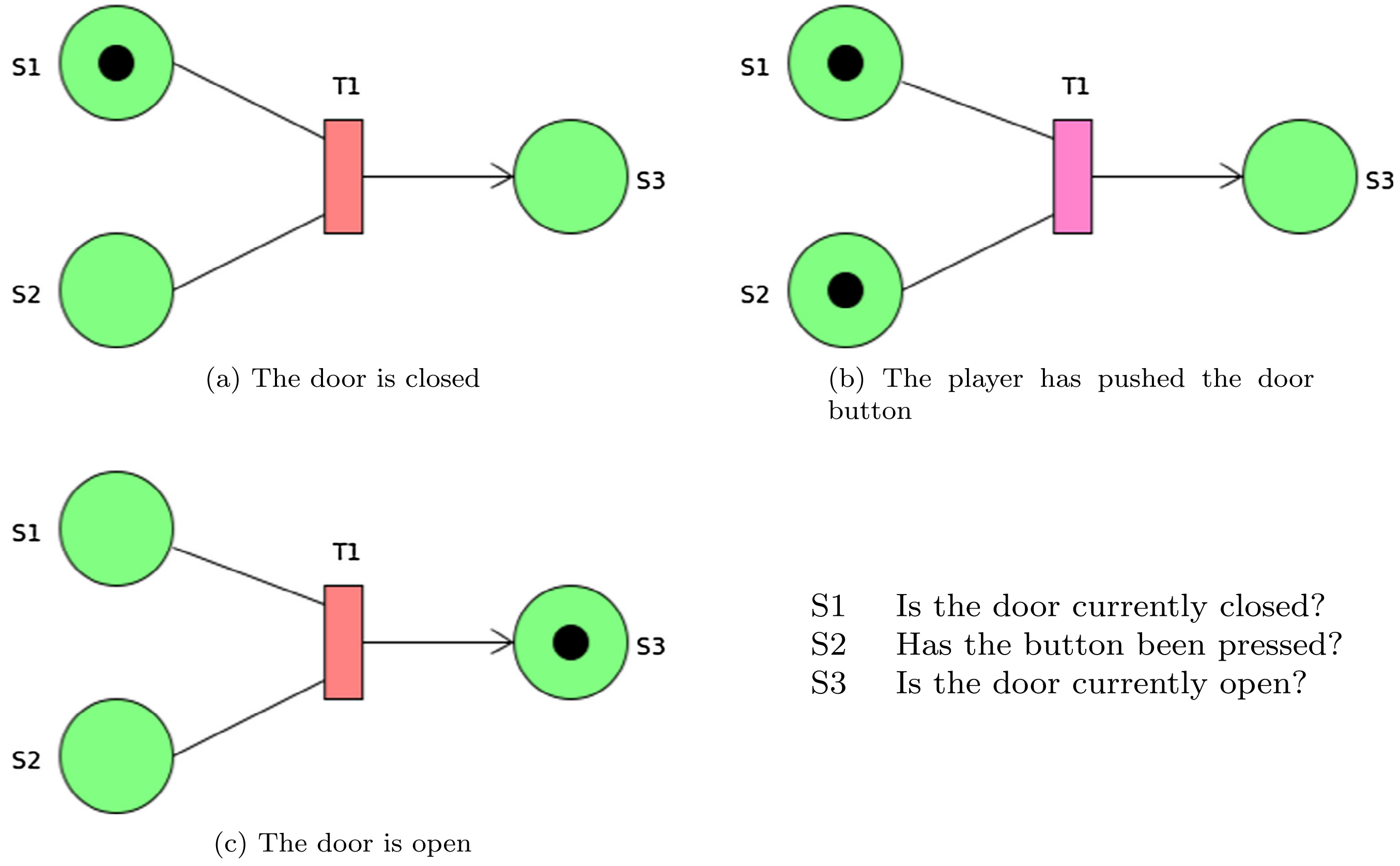

**Fig. 6.** Modelling the action of opening a door with a Petri net.

Our proposal is to model games using Petri nets to adapt and execute *high level actions*, mentioned previously in Section 5.3 and saved in the traces during the recording of game sessions played by beta testers. As specified before, the sequence of high level actions recorded are a reliable source of information of the steps that the player carried out. Our objective is to use them to check that the playability of the game is still valid even when the level has been modified.

When the "high level tests" are replayed, the raw input generated by the player is not important anymore, so it is simply discarded. What we use now are the high level messages generated, which we stored conveniently. Those messages are directly related to the specific transitions from the Petri net that models the game. Every time a transition of the net is fired, we could say that the game "evolves".

At the time of replay of a "high level test", we can see that at a certain point a message was generated (for example, "Open door"), and the transition associated to this message in the Petri net tells us that in order for this message to be produced in the original execution and the transition to be fired, several preconditions had to be fulfiled at the time: the door had to be closed and the player should push the button. The reproduction of the "high level test" will handle trying to fulfil the preconditions so that the original message can be produced again.

It is clear that to achieve all this we need to have a good understanding of the domain first; that is, of the game itself. In particular, we need to define the Petri net that models the behaviour of the game, as well as the relation between each transition and its "high level message", and between each place and the precondition (state of the game) that it models.

Furthermore, we need to bridge the gap between preconditions and how to fulfil them. After all, the replay of high level tests can imitate how a human plays and act upon the actions of the player avatar as if these movements were the ones performed by the human player; the state of the game cannot be modified under the hood. Therefore, it is necessary to know, for instance, that to fulfil the precondition "the player pushed a button" associated to a particular place in the Petri net, we have to move the player close to the button, which is equivalent to having a token on that place in the net.

The following section presents an example of use of all these ideas in a real game, where we explain some of the tests we ran for it and show a small example of the traces recorded when playing the game.

## 8. Example of use: Time and Space

*Time and Space* is a puzzle platformer developed by a group of students of *the Máster en Desarrollo de Videojuegos* ("Master in Videogame Development"). The game is organised in several levels, being the objective of each of them to get to a light portal placed somewhere in the map of the level. The human player controls only the main player character, directed in a first-person perspective.

What is different about *Time and Space* is the way in which all the different levels can be completed. The player can decide to make a copy of himself anytime during the duration of the level, causing it to restart and materialising not only the main player now, but also all the clones created. Fig. 7 shows a screenshot of the game where we can see the main player standing next to one of its copies. These copies constitute the main game mechanic, and are useful because they *replicate the movements* of the main character in previous executions, helping him reach his objectives.[2]

To illustrate this, we are going to describe level 1 of the game (Fig. 8). To be able to reach the light portal at the end of the level, the main player has to walk through two different doors. But the player cannot do this alone. To open each of these doors it is necessary to stand on a switch that is far away from the door. If the player moves from the switch to a different place, the door will close again, so he needs to create clones that will help him keep these doors open. Having said this, the actions needed to reach the end of level 1 go as follows: the main player has to stand on the first switch and then clone himself. After that, when the level is restarted again, the clone will behave exactly as the main character did before and it will stand on the switch, keeping the first door open. Now, the main player, directed by the human player, can walk through the first door and stand on the other switch to open the second door. This is the exact moment that is shown in Fig. 8. Making a second clone of the player will restart the level again, and now each of the clones will keep one of the doors open so that the main player can walk towards the light portal, reaching the final destination and finishing the level.

The recording of the traces for testing purposes is made on a modified version of the game that dumps onto the disk the raw input detected and the most important internal messages.

The raw input is logged to a file with a very simple format. Each of the lines in this file corresponds to an event produced with the input devices (in our case, keyboard and mouse), and it contains several fields: the input device, the key pressed/unpressed or the mouse movement, and the timestamp of the event.

On the contrary, the saved information about the internal messages is richer and more complete. For the serialisation of these messages we use JavaScript Object Notation (JSON) because of its simplicity and versatility, as it allows to have both named objects and arrays. An example of a "high level trace" (i.e. internal messages) recorded when playing *Time and Space* is shown in Fig. 9. This message was generated when a button (`Button1`) was pressed, and asked its associated door (`Door1`) to open.

For running our tests, we first recorded game traces from level 1 of *Time and Space*, without modifying the level map at all. Then, we reproduced the generated raw input traces from the user to check that both the recording and replaying systems were working in the game. The result of our test was successful and the execution reached the end of the level.

In order to be able to execute a *high-level test*, we need to model the game dynamics using a Petri net. *Time and Space* doors are slightly different from that ones in Fig. 6 because the mechanism for opening them is not equivalent. Whilst in the first example pushing the button causes the door to open immediately, in *Time and Space* they behave differently and they only remain open for as long as the player (or a clone) is standing on the switch, going back to the original position when the player moves away from it. Taking this into consideration, Fig. 10 shows the Petri net for a door in *Time and Space*. Places in the net model game states, whilst transitions model *events* that make the game evolve. They occur automatically in the game simulation. It should be noted that $(S_1, S_3)$ and $(S_2, S_4)$ are pairs of opposite states, but this is not a problem for the model and its use during trace playing.

The first level of *Time and Space* contains two of these doors, so its model has the Petri net replicated. "High level traces" stored during the human tester execution are nothing more than the underlying messages exchanged between components and entities. For the execution of the *high-level tests*, only some of them are important, and are listed in Table 1. They are the messages that make the game evolve, advancing through the resolution of the puzzle. It is not a coincidence that OPEN messages correspond with transitions $(T_1)$ in the Petri net of Fig. 10.

After defining the model of the level with a Petri net, we tried to run different tests to see if everything was working as expected. First, we modified the map and we moved the switches to different places, but we made sure that the new position was still reachable from the point of view of the player. After this, we ran our test and we could observe that the Petri net was checking what

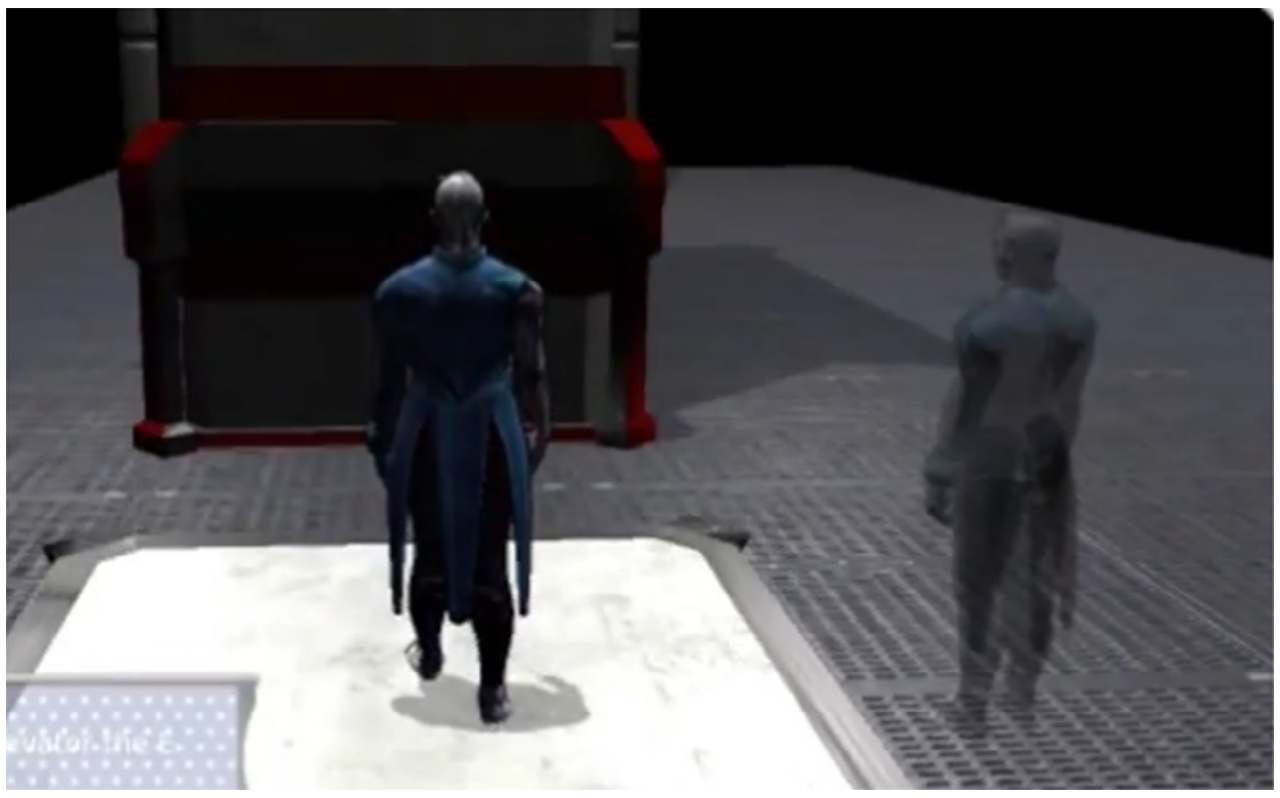

**Fig. 7.** Screenshot of *Time and Space*.

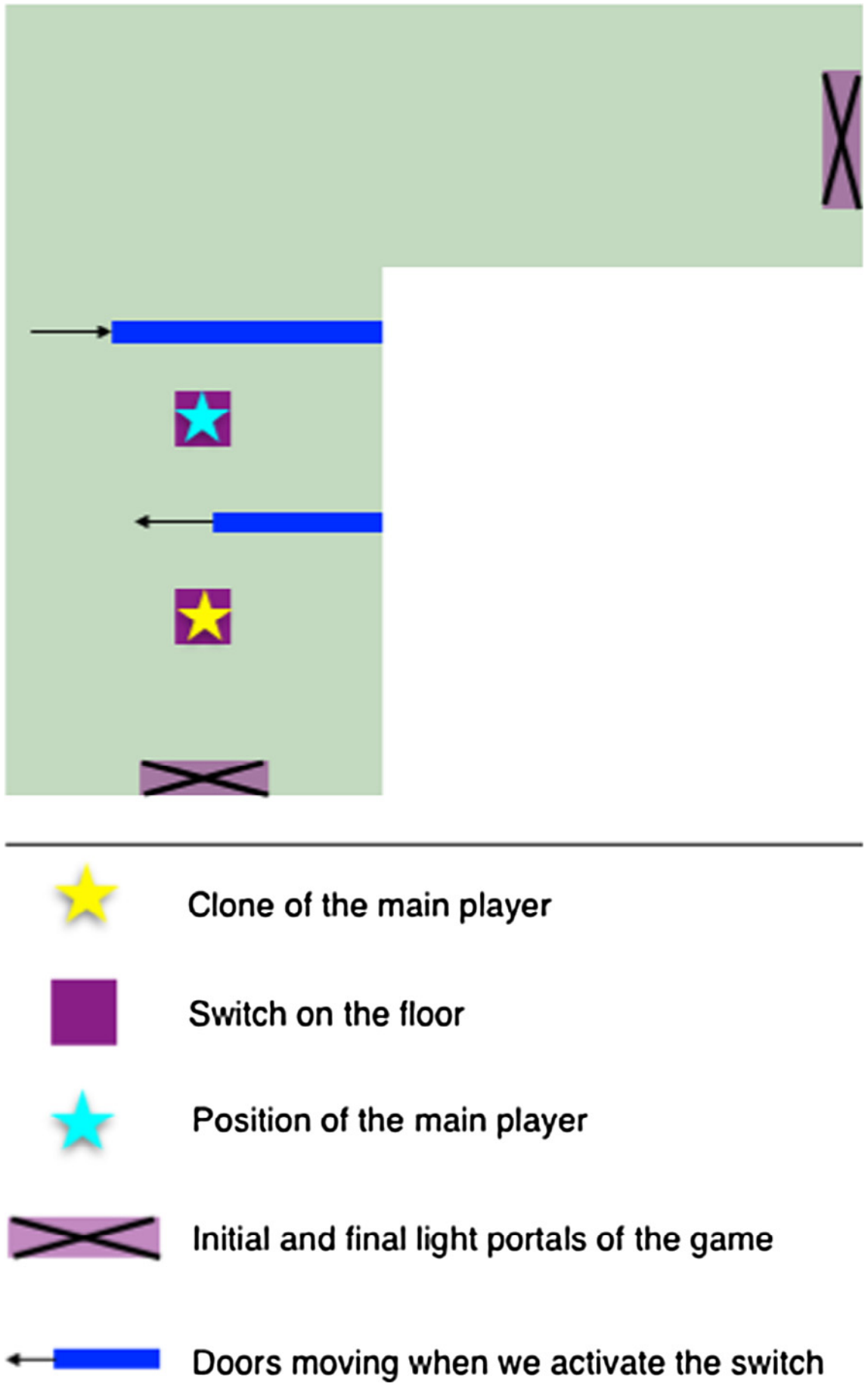

**Fig. 8.** Outline of level 1 of *Time and Space*.

[2] Full gameplay shown at https://www.youtube.com/watch?v=GmxV_GNY72w.

```
{
  "timestamp" : 73400,
  "type" : "Open",
  "SourceEntity" : {
    "name" : "Button1",
    "type" : "DoorButton"
  },
  "TargetEntity" : {
    "name" : "Door1",
    "type" : "Door"
  }
}
```

**Fig. 9.** Example of a trace generated when a button requests a door to open.

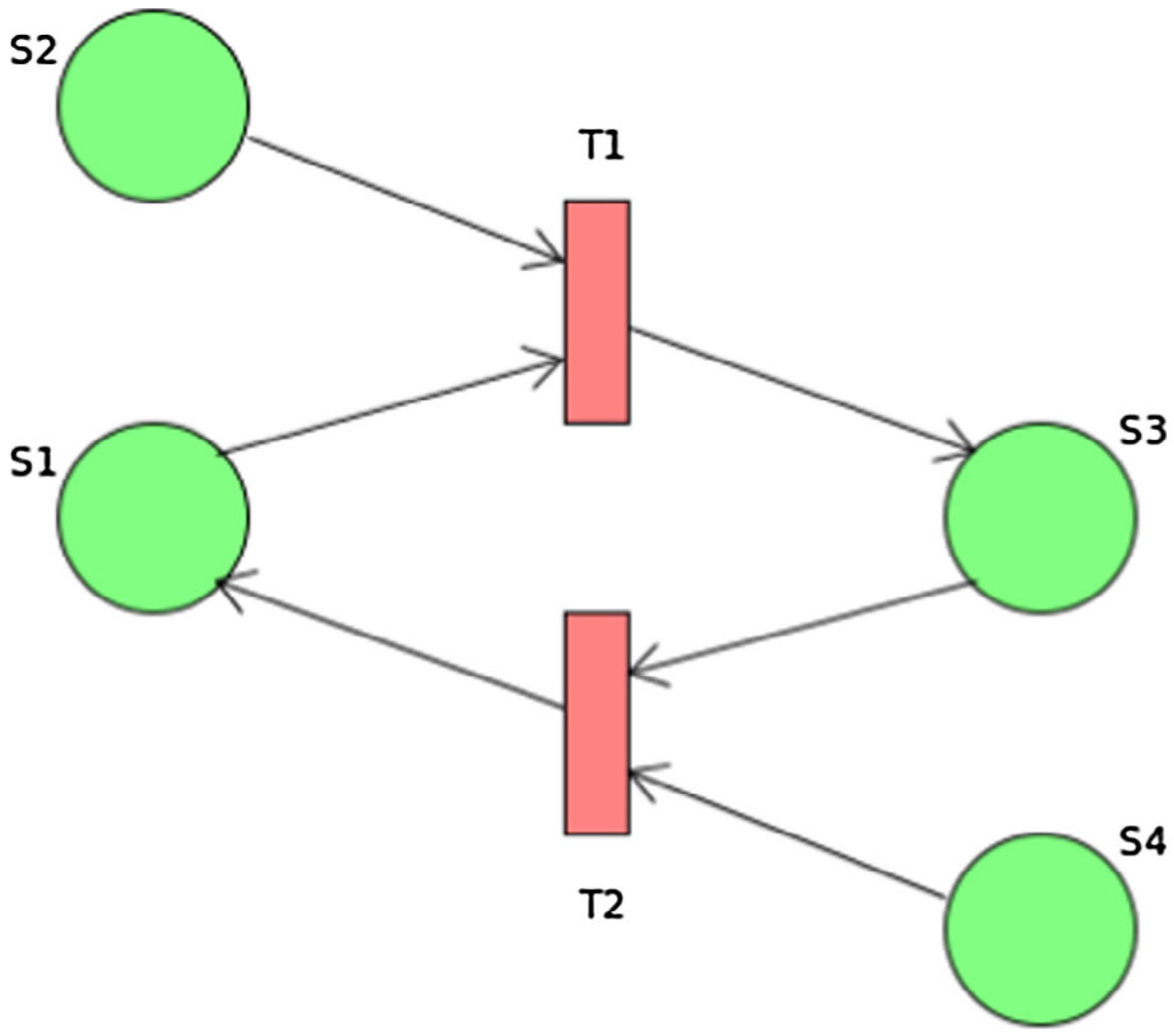

| | |
|---|---|
| S1 | Is the door currently closed? |
| S2 | Is the button currently being pressed? |
| T1 | Button requests the door to open |
| S3 | Is the door currently open? |
| S4 | Is the button currently unlocked? |
| T2 | Button requests the door to close |

**Fig. 10.** Petri net for doors in *Time and Space*.

**Table 1**
High level trace.

| Source entity | Type | Target entity |
|---|---|---|
| Button 1 | Open | Door 1 |
| Player | Clone | – |
| Button 1 | Open | Door 1 |
| Button 2 | Open | Door 2 |
| Player | Clone | – |
| Button 1 | Open | Door 1 |
| Button 2 | Open | Door 2 |
| Player | Touch | End portal |

preconditions were fulfiled and which were not. In the execution, the movement of the player avatar was adapted thanks to the AI of the game and it was possible to reach the end of the level.

After this experiment, we modified the level to run a second test. Now, we modified the position of one of the switches and we placed it behind a closed door, making it impossible for the game to advance. Again, the Petri net allowed us to check on run-time what preconditions were not fulfiled, so the test ended in a failure.

After the success of these two tests, we decided to try out our system in a completely different level of the game, where performing more complicated actions was required to reach the light portal. We chose level 4, in which the player needs to walk through a platform that is moving, has to deactivate some rays that would kill him otherwise and, moreover, needs three clones instead of two to help him reach the end of level mark. The execution of the test, once again, was successful and ran as expected.[3]

## 9. Implementation requirements

In order to incorporate into a videogame the ideas proposed in this paper, it will need to be executed in two new different modes, apart from the basic execution used by players:

- *Recording mode*: beta testers will use this mode when recording traces. The game will dump into a file the *raw user input*, and the internal messages between components that programmers have considered important enough.
- *Testing mode*: it will be used for *running tests*. In this mode, the original trace recorded must be received as an input, as well as the necessary configuration as described in Section 6. The game will be executed autonomously, either re-injecting without any modifications the original input or adjusting the execution (*high level tests*).

[3] A reproduction of these tests is shown in https://www.youtube.com/watch?v=1OBlBKly1pk.

In both of the cases, we will avoid showing the typical initial menus of the game and merely launch a level directly.

In order to include these new modes in a game, the code of our game has to be modified. Before going into these changes in the following subsections, it is worth mentioning the need for the game to perform a simulation with *fixed time steps* so that the execution is *repeatable*.

Essentially, a videogame is just an infinite loop where the simulation advances first, and then graphics are drawn. Depending on the speed of the computer, existing tasks in the background and even having the power-save mode active, the number of times per second that the game will loop is different. This means that, unless we take measures to avoid it, in each execution of the game the time instants in which we will simulate the state of the game (*game time*) will be very different. This *ruins completely* the possibility of re-running the execution, considering that very small differences can make the result to vary quite a lot.

To solve this, it is required to decouple the real computer speed, and hence the number of times running the main loop, from the moments where the state of the world is simulated. When this is achieved, it is said that the game is using a *fixed time step loop* [20,7,10]. Fortunately, nowadays the vast majority of games implement this kind of loop because of its multiple benefits, so this restriction is not too harmful.

### 9.1. Recording mode

In this mode of execution, the game has to dump both the data related to the raw traces and the internal messages from the components that are considered important onto the disk. To do that, knowing the internal structure of the code of the game is necessary.

Thanks to the use of component-based architectures (Section 3), the process of saving the messages is very straightforward.

Specifically, it is sufficient to create a new component, called `CRecorder` in our testbed game, *Time and Space*, and associate it to different entities instead of modifying the main code of the game. Using a new component has the following advantages:

- Virtually all videogames using components are *data-driven* and they store what components are part of each entity in external files. Adding a new component in different entities is fast and does not require recompiling the code.
- Depending on the objectives of each of the tests, it can happen that we are not interested in all the messages generated by some of the entities. For this reason, it is not necessary to register `CRecorder` as a component in every single entity of the game. By avoiding doing this, more precise logs are generated. Having representative logs is, as mentioned in Section 5.2, very important to programmers due to the fact that they can easily identify what happened during the execution of the gameplay.
- Not all the messages generated by an entity during a gameplay are equally important. Because we have our new component `CRecorder` registered for specific entities, we can design a filter of all the type of messages that we want to be listeners of and save to the log only traces of those that are interesting enough for the programmer and to be replayed later.
- Because the testing system is only related to the main code of the game through the component system, it is possible to keep developing and changing the code of the game without causing any trouble. When the testing phase ends, removing everything related to the tests is as simple as deleting the name of the recording component from each of the entities. At this point, the code of the game and the testing system are not related anymore.

For instance, let us consider an example in which a player shoots an enemy. In this situation, it is possible that the physics system stops working somehow and the enemy does not receive any damage even if the bullet hits him. To keep track of this, we can register two different recorders in the component system: one associated to the player entity, and the other one to the enemy entity. The first recorder will save the messages generated when the player shoots, and the second one the messages produced when the bullet hits the enemy. By doing this, the final log will not contain useless information, and if the test fails, the programmer will be able to easily find out where the error is.

`CRecorder` allows us to save all the internal messages that we need, but we still have to figure out how to log raw inputs from the users. This process can normally be done by adding specific code in the input manager to serialise the received information, together with the *timestamp*.

A different option that may be more convenient is to integrate the recording of this information with the recording of the messages described before. Videogames often have a low level input manager that process raw input and turns it into messages directed to the player avatar entity, such as "walk", "jump" or "shoot". Using this, the recording of the input becomes integrated with the recording of the messages, reducing the amount of code that is necessary to add and, even better, simplifying also the so-called injection of "raw input" when we execute tests. At this time, it is not necessary to "simulate" input devices, since sending the messages to the entity is enough.

As an example, the code for the `CRecorder` component and the required changes made in the original implementation of *Time and Space* game were about 8 KB.

### 9.2. Testing mode

Executing the game in a testing mode requires that we can run the game in a completely unattended and autonomous way. Moreover, having direct access to the messages system to inject them (to the player avatar) and monitor them is required. We can use a slightly modified version of the `CRecorder` component for this purpose.

However, the most complex part of executing the game in this mode is related to the adaptation of the traces for running high-level tests.

As described in Section 7, we propose the use of Petri nets to adapt traces recorded previously and to decide which actions to reproduce at each time of the gameplay.

Our aim is to replay the level using just the *high-level trace* summarised in Table 1. Note that this trace contains the level *walkthrough*; a human player could use it as-is for completing the level. Our use of Petri nets will allow us to use this trace to automatically replay the level even if the map has changed. If the level can be completed, we will have proved that the gameplay is still valid and the changes have not broken the level.

When the test executor finds the first message, "Button 1 Opens `Door 1`" it must be able to relate it with the transition $T_1$ of the Petri net. Once done, it is easy to go backwards and figure out the preconditions, $S_1$ and $S_2$, that are related to the game state "the door is closed", which is true, and "the button is currently pressed", which is not.

The test executor wants $T_1$ to be fired (in order for the message in the trace to be sent) so it will try to make $S_2$ become true. That is accomplished by moving the player avatar to the button position. Once there, the game simulation will automatically send the message and the door will open. The executor will detect that message thanks to the modified `CRecorder` component, and it will advance the execution to the next message. Note that the Petri net points out that, after that, $S_3$ will have a token. Although we are not currently doing it, the test executor could confirm this by looking at the game state associated with it (the door is now open).

In order for us to use Petri nets, three aspects have to be resolved first:

- Petri net representation that models the game: we need to represent the net somehow, defining places, transitions and arcs.
- Relation between the Petri net and the game:
  - In the formal representation of the game, transitions of the net are those that, when fired, make the game advance. At the same time, in the actual game, execution advances because of the messages sent between components. The high-level test executor needs to bridge this gap by learning the relation between the messages in the trace and the transitions that model the game as a Petri net, for example, knowing that the transition $T_1$ has been fired when `Button1` sends an Open message to `Door1`.
  - On the other hand, Petri net places (and its tokens) indicate which preconditions have to be fulfiled in order to fire a transition. These preconditions are *game states* ("the door is open"), so the test executor needs to know how to relate the state of the game to the tokens on places.
- For all those places representing preconditions that can be accomplished directly by the player ("The button is pressed"), it is necessary to know how to manipulate the player avatar so that these preconditions are satisfied. This often means having access to the player avatar navigation features.

Having a explicit representation of a Petri network is a relatively easy task; they can be implemented using a graph as a starting point and adding tokens and transitions becomes just a data structure exercise. Storing the relationship between transitions and messages between components is also fairly easy because messages are *data* by themselves, so it is not hard to associate a transition in the net with a specific kind if message. Nevertheless, care must be taken for keeping the information regarding the *concrete map entities*. For example, the level used in our experiment had two instances of the Petri Net in Fig. 10, each one referring to a different door and button. When creating the explicit model for the level, the links between transitions and their entities must be created.

A more challenging task is to get a declarative representation of the relationship between places in the Petri net and states in the game. In *Time and Space*, the state of the game is not stored during runtime in such a way that allows accessing the state in a generic way (with a method like `bool getState("isOpen")`). Although some games provide this feature (useful to export the state to script languages), our test bed game lacks of it. In our experiment, the number of places in the net is quite small, so instead of overengineering the source code by adding this feature, we decided to implement a specific test class for each place in the net. Although this spoils the genericity of our implementation, we consider that a generic implementation is in fact possible so the technique can be extensively used, not as in our experimental implementation.

The situation is quite similar regarding the relationship between places and how the precondition they model can be satisfied using the avatar ($S_2$, is the button pressed?, in our experiment). A generic implementation would require an avatar player controller that could receive declarative commands such as "Go to Button1". If that was the case, we could declaratively link those places with game commands to be launched. Again, *Time and Space* does not provide this feature, and a hand-written implementation has been created.

Even if the solution is almost exact, this method has some limitations. Using Petri nets to reproduce traces like that requires that the videogame has an Artificial Intelligence (*AI*) implemented, capable of controlling the player inside the game. Fortunately, as we said before, there are a lot of games (like *Time and Space*) that use an AI for directing all non-player characters' movements, which can be reused for the replay of the game. However, despite the fact that the results we have from *Time and Space* are very promising, there are some use cases in which the reproduction of traces does not work properly. One of these examples is when the player needs to push a button that is not on the same level as the one in which the player is standing. In this case, the AI of the videogame cannot find out where the button is, so the reaction of the player will be to just wait there without moving. This situation shows that the testing method proposed is valid, but remarks that a proper AI designed to control the player is needed.

If we do not have a proper AI and there is no chance to develop one, a way to try to avoid this situation is to combine both ways of running tests described in Section 6. When an action that the AI cannot reproduce is detected, instead of reporting that the test failed, we can switch from the method that uses the AI to reproduce the traces to the one that injects raw input into the system. This way, we can succeed in fulfiling the test requirements and make the tests pass. For example, if we have to jump into an elevator to go to an upper floor and the AI does not know how to do it, we can just inject the key that handles the "JUMP" action when we are in front of the elevator and then just continue reproducing from there.

To finish this section, it is worth mentioning that for executing our tests we considered the possibility of using an "accelerated mode". This mode would consist on removing all the graphical components, and thus speeding up the time needed to carry out the tests. Thanks to the component-based architectures, this specialised execution does not require a lot of additional development work.

However, even if this might make sense for some of the cases, removing components in the execution can lead to running less precise tests. In particular, an execution in which we do not test rendering modules cannot be accepted as valid for compatibility and regression tests. Nevertheless, it could be appropriate to run tests focused towards checking the playability of the game after changing a map level rather than testing the source code.

## 10. Conclusions and future work

Videogames are software programs composed of a myriad of lines of code that can, and usually do, have bugs. Classic techniques, as unit testing, can be used to detect them. Unfortunately, videogames are not just software: they are also composed of *assets* such as 3D models, textures, or music that provide the *art* and, more important, maps and puzzles that provide the *gameplay*. During a videogame production, the base code evolves, and so does the gameplay, but classic *software testing* is not useful for these assets.

This paper presented an approach to *beta testing* that is useful for running compatibility and regression tests and, at the same time, for checking that the gameplay is still correct after level changes. Taking advantage of the component-based architecture, game session traces are recorded in order to be replayed (and maybe adapted) in subsequent game versions for checking their correctness. This adaptation is based on the use of Petri nets for modelling the game dynamic.

The proposal has been tested in a small game, *Time and Space*, with promising results. The Petri net has an explicit representation on runtime, and the relationship between component messages and their corresponding transitions in the Petri net is also data driven, so they could be used in other games with only minor changes. On the contrary, the union between places in the net and states in the game is hard-coded. Although it could be implemented in a more generic way, it would require some support in the entities and components subsystem that the *Time and Space* implementation does not provide.

A natural evolution of this work would be to test all these ideas in other games. An easy way to achieve this goal would be to migrate the implementation to a component-based game engine such as Unity3D. Developing games over it is easier than creating them from scratch, and once implemented, the testing framework would be reused and polished again and again. The implementation would have to overcome some technical challenges regarding the access to the low level input traces. It would also impose programmers some kind of adaptation because although Unity is component based, messages can be avoided and plain method invocation is usually used, a habit that ruins our technique.

## Acknowledgements

This work was funded in part by the Spanish Ministry of Economy and Competitiveness (TIN2014-55006-R) and Madrid Education Council and UCM (Group 910494).

## References

[1] I. UK, GTA 5 Makes $800 Million in One Day, 2013 <http://uk.ign.com/articles/2013/09/18/gta-5-makes-800-million-in-one-day> (accessed July, 2016).
[2] Wikipedia, Development of Grand Theft Auto v – Wikipedia, The Free Encyclopedia, 2015 <https://en.wikipedia.org/w/index.php?title=Development_of_Grand_Theft_Auto_V&oldid=696260870> (accessed July, 2016).
[3] L. Mellon, Automatic Testing for Online Games, 2006.
[4] S. Gandini, W. Ruzzarin, E. Sanchez, G. Squillero, A. Tonda, A framework for automated detection of power-related software errors in industrial verification processes, J. Electron. Test. 26 (6) (2010) 689–697.
[5] M. Mehta, S. Ontañón, T. Amundsen, A. Ram, Authoring behaviors for games using learning from demostration, in: Case-Based Reasoning for Computer Games Workshop.
[6] S. Ontañón, K. Bonnette, P. Mahindrakar, M.A. Gómez-Martín, K. Long, J. Radhakrishnan, R. Shah, A. Ram, Learning form human demonstrations for real-time case-based planning, in: STRUCK-09 Workshop.
[7] J. Gregory, Game Engine Architecture, second ed., A.K. Peters, Ltd./CRC Press, 2014.
[8] Wikipedia, Half-life (video game) — Wikipedia, the Free Encyclopedia, 2015 <https://en.wikipedia.org/w/index.php?title=Half-Life_(video_game)&oldid=695654947> (accessed July, 2016).
[9] S. Bilas, Lecture in Game Developers Conference, March 2002 <http://scottbilas.com/files/2002/gdc_san_jose/game_objects_slides_with_notes.pdf>.
[10] B. Nystrom, Game Programming Patterns, Genever Benning, 2014.
[11] E. Gamma, R. Helm, R. Johnson, J. Vlissides, Design Patterns: Elements of Reusable Object-oriented Software, Addison-Wesley Longman Publishing Co., Inc., Boston, MA, USA, 1995.
[12] E. Folmer, Component based game development: a solution to escalating costs and expanding deadlines?, in: Proceedings of the 10th International Conference on Component-based Software Engineering, CBSE'07, Springer-Verlag, Berlin, Heidelberg, 2007, pp 66–73. <http://dl.acm.org/citation.cfm?id=1770657.1770663>.
[13] Software Engineering Technical Committee of the IEEE Computer Society: IEEE Std 829-1998, IEEE-SA Standard Board, 1998.
[14] G. Meszaros, XUnit Test Patterns: Refactoring Test Code, Addison-Wesley, 2007.
[15] Lionbridge, Lionbridge Reports First Quarter Results with Record Revenue, 2015 <http://www.lionbridge.com/lionbridge-reports-first-quarter-2015-results/> (accessed July, 2016).
[16] L. Popova-Zeugmann, Time and Petri Nets, Springer-Verlag, Berlin Heidelberg, 2013.
[17] M. Estevão Araújo, L. Roque, Modeling games with Petri nets, in: DIGRA2009 - Breaking New Ground: Innovation in Games, Play, Practice and Theory.
[18] D. Kafura, Notes on Petri Nets, 2011 <http://people.cs.vt.edu/kafura/ComputationalThinking/Class-Notes/Petri-Net-Notes-Expanded.pdf> (accessed July, 2016).
[19] K. Jensen, An Introduction to the Practical Use of Coloured Petri Nets, 1998 <ftp://calhau.dca.fee.unicamp.br/pub/docs/gudwin/Petri/use.pdf> (accessed July, 2016).
[20] D. Sánchez-Crespo, Core Techniques and Algorithms in Game Programming, New Riders Games, 2003.